\documentclass[lettersize,journal]{IEEEtran}
\usepackage{amsmath,amsfonts}
\usepackage{algorithmic}
\usepackage{algorithm}
\usepackage{array}
\usepackage[caption=false,font=normalsize,labelfont=sf,textfont=sf]{subfig}
\usepackage{textcomp}
\usepackage{multirow}
\usepackage{stfloats}
\usepackage{url}
\usepackage{verbatim}
\usepackage{graphicx}
\usepackage{amsmath}
\usepackage{amssymb}
\usepackage{booktabs}
\usepackage{xcolor}
\usepackage{color, colortbl}
\usepackage{verbatim}
\usepackage{cite}
\usepackage[numbers,sort&compress]{natbib}
\usepackage[pagebackref,breaklinks,colorlinks]{hyperref}
\usepackage[capitalize]{cleveref}
\crefname{section}{Sec.}{Secs.}
\Crefname{section}{Section}{Sections}
\Crefname{table}{Table}{Tables}
\crefname{table}{Tab.}{Tabs.}

\newcommand{\mypara}[1]{\vspace{1mm}\noindent\textbf{#1}}

\begin{document}

\title{MonoGAE: Roadside Monocular 3D Object Detection with Ground-Aware Embeddings}

\author{
Lei Yang\textsuperscript{1}, Jiaxin Yu\textsuperscript{2},  Xinyu Zhang\textsuperscript{1}, Jun Li\textsuperscript{1}, Li Wang\textsuperscript{1}, Yi Huang\textsuperscript{1}, Chuang Zhang\textsuperscript{1}, Hong Wang\textsuperscript{1}, Yiming Li\textsuperscript{3}\\

\textsuperscript{1} School of Vehicle and Mobility,  Tsinghua University \\
\textsuperscript{2}South China University of Technology;
 \textsuperscript{3}New York University \\
\begin{normalsize}${\tt \{yanglei20, huangyi21, zhch20\}@mails.tsinghua.edu.cn; 202121010334@mail.scut.edu.cn}$\end{normalsize} \\
\begin{normalsize}${\tt \{xyzhang, lijun1958, wangli\_thu@mail\}.tsinghua.edu.cn};$\end{normalsize} 
\begin{normalsize}${\tt yimingli@nyu.edu}$\end{normalsize}}

\markboth{Journal of \LaTeX\ Class Files,~Vol.~14, No.~8, August~2021}%
{Shell \MakeLowercase{\textit{et al.}}: MonoGAE: Monocular Roadside 3D Object Detection with Ground-Aware Embedding}


\maketitle

\begin{abstract}
Although the majority of recent autonomous driving systems concentrate on developing perception methods based on ego-vehicle sensors, there is an overlooked alternative approach that involves leveraging intelligent roadside cameras to help extend the ego-vehicle perception ability beyond the visual range. We discover that most existing monocular 3D object detectors rely on the ego-vehicle prior assumption that the optical axis of the camera is parallel to the ground. However, the roadside camera is installed on a pole with a pitched angle, which makes the existing methods not optimal for roadside scenes. In this paper, we introduce a novel framework for Roadside Monocular 3D object detection with ground-aware embeddings, named MonoGAE. Specifically, the ground plane is a stable and strong prior knowledge due to the fixed installation of cameras in roadside scenarios. In order to reduce the domain gap between the ground geometry information and high-dimensional image features, we employ a supervised training paradigm with a ground plane to predict high-dimensional ground-aware embeddings. These embeddings are subsequently integrated with image features through cross-attention mechanisms. Furthermore, to improve the detector's robustness to the divergences in cameras' installation poses, we replace the ground plane depth map with a novel pixel-level refined ground plane equation map. Our approach demonstrates a substantial performance advantage over all previous monocular 3D object detectors on widely recognized 3D detection benchmarks for roadside cameras. The code and pre-trained models will be released soon.
\end{abstract}

\section{Introduction}
\begin{figure}[!t]
\centering
\includegraphics[width=0.49\textwidth]{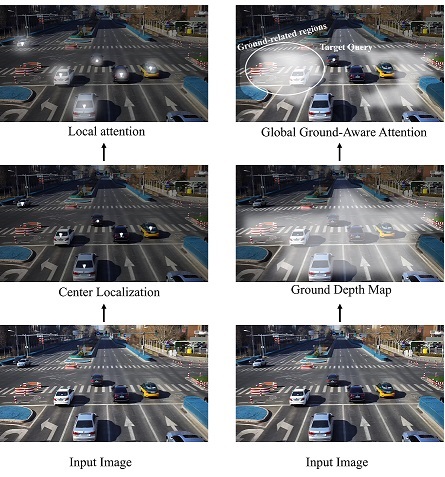}
\caption{\textbf{Center-based Pipeline v.s. Ground-aware Pipeline.} Traditional Center-guided Methods (Left) and our Ground-guided Paradigms (Right). Existing center-guided methods utilize features around the centers to predict 3D attributes of objects, while our method guides the whole process through the predicted ground information and adaptively aggregating ground features from the entire scene. The upper right picture denotes the attention map from the ground cross-attention layer.}
\label{fig_1}
\end{figure}
\IEEEPARstart{M}{onocular} 3D object detection is the task of estimating three-dimensional information solely from a single 2D image, offering extensive applications in real-world scenarios, including autonomous driving and robotics. Due to its low cost and closer proximity to mass production, it has attracted increasing attention from researchers in academia and industry. However, existing research has mainly focused on ego-vehicle applications \cite{10061347,9780191,9729810}, where the camera's position is close to the ground and obstacles can be easily occluded by other vehicles. This greatly limits the ego-vehicle perception capabilities and further leads to potential safety hazards in autonomous driving. Therefore, researchers have begun studying roadside perception systems using higher-mounted intelligent sensors, such as cameras, to solve this occlusion problem, expand the perception range, increase the reaction time for autonomous driving in dangerous situations through cooperative techniques \cite{yu2023vehicle,fan2023quest,10077757,10122468,9228884}, and thereby improve safety. In order to promote future research, some large-scale roadside datasets \cite{yu2022dair, ye2022rope3d, yu2023v2x} containing images collected from roadside view and corresponding 3D annotations, have been released to provide an important basis for training and evaluating roadside monocular 3D object detection methods.

Roadside data has significant characteristics compared to on-board scenes. Firstly, the camera is positioned higher and captures a top-down view of the road scene, providing a wider field of view and observing smaller and more numerous objects. Secondly, the background of roadside images is usually a fixed road with strong prior information. Based on the depth of an object's position on the ground, its depth in 3D space can be determined. By using the provided ground equation, pinhole model, and camera pose, the depth of each pixel corresponding to the ground can be derived. Therefore, directly applying traditional on-board monocular 3D object detection methods \cite{rukhovich2022imvoxelnet,brazil2019m3d} to roadside scenes is not the optimal choice, as shown in \cref{fig_1}~(left). To achieve the best roadside monocular 3D object detection, the key is to reasonably utilize the strong prior information of the fixed road surface.
\IEEEpubidadjcol

In order to combine ground geometry information with high-dimensional semantic features to achieve accurate monocular 3D object detection on the roadside senarios. There are two main technical challenges: first, there exists a significant domain gap bettween the ground geometry information and high-dimensional image features, requiring effective fusion techinique. Second, due to the diversity of camera installation poses, the ground geometry information in the camera coordinate system varies significantly across different road scenes. Therefore, it is necessary to choose an appropriate ground information encoding techique to improve the generalization performance of detectors from known to unknown scenes.

\begin{figure*}[!t]
\centering
\includegraphics[width=7in]{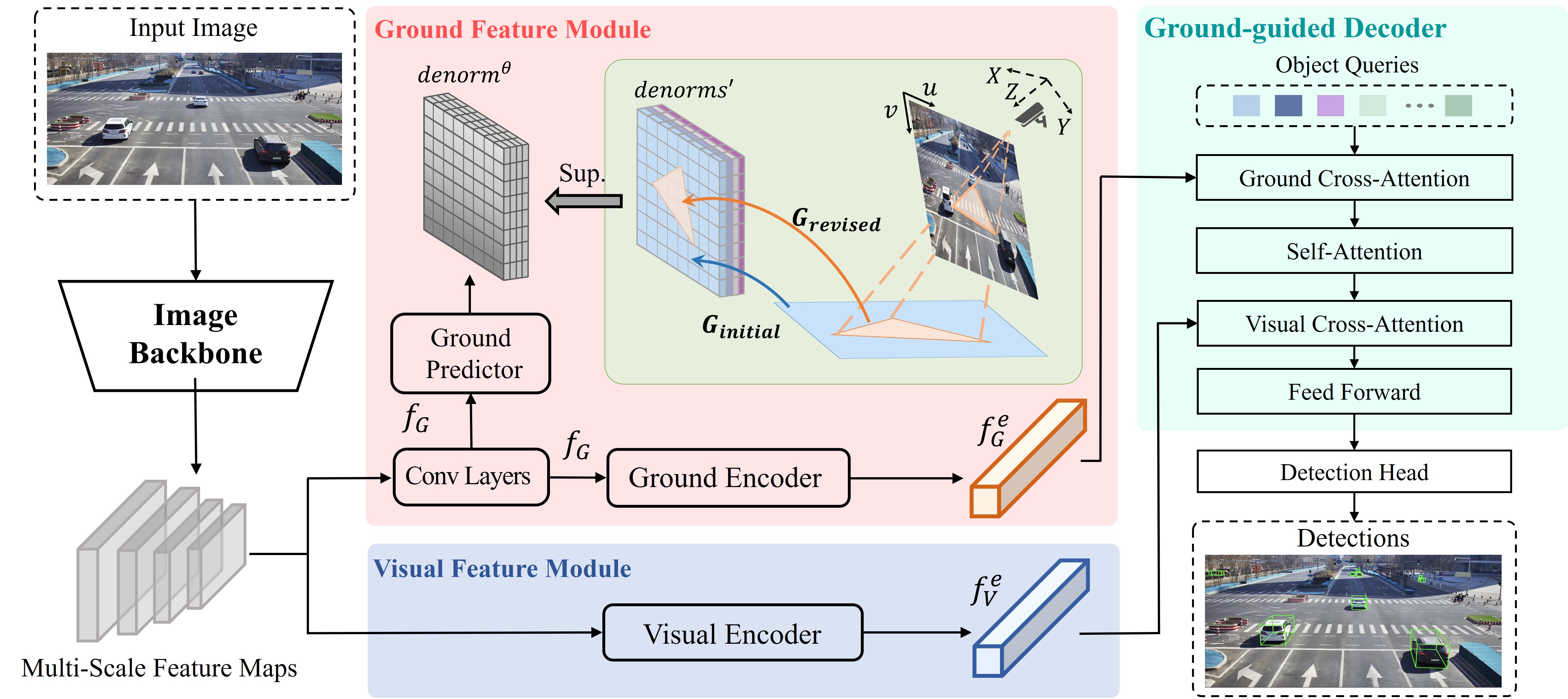}
\caption{\textbf{The overall framework of MonoGAE.} First, image backbone extracts high-dimensional image features. Then, image features are fed to the Ground Feature Module to generate high-level ground-aware features. Visual features are produced by the Visual Feature Module with the same image features. Ground-guided Decoder fusion both ground-aware features and visual features though cross-attention to produce the final predictions.}
\label{fig_2}
\end{figure*}

In this paper, we proposes a roadside monocular 3D object detection framework based on ground-aware embedding, named MonoGAE (Monocular 3D Object Detection with Ground-Aware Embedding). Unlike existing methods that directly fuse ground geometry information and high-dimensional semantic features, we adopt a supervised training paradigm. During the training phase, we use ground geometry information as the ground truth of the Ground Predictor to guide the model in generating high-dimensional features that encode implicit ground-aware features, as shown in \cref{fig_1}~(right). This enables the mapping of ground geometry information and high-dimensional semantic features to a common feature space. In the inference phase, we apply a ground-guided decoder to fuse the implicit ground-aware high-dimensional features and the high-dimensional semantic features, to estimate the 3D attributes of each object globally, as shown in \cref{fig_2}. The MonoGAE framework consists of three core modules: the Ground Feature Module (GFM), the Visual Feature Module (VFM), and the Ground-guided Decoder. VFM is responsible for generating high-dimensional semantic features, GFM generates implicit ground-aware high-dimensional information through auxiliary task supervisory training, and the Ground-guided Decoder fuses high-dimensional semantic features and implicit ground-aware high-dimensional features from VFM and GFM using a cross-attention mechanism.

To address the challenges of generalization and robustness caused by the diversity of camera installation poses in roadside scenarios, we proposes a pixel-level ground plane equation map encoding method as the ground truth for ground predictor. Compared to the depth map of ground , this has significant robustness and improves generalization.

We conducted extensive experiments to validate the effectiveness of the proposed method. In terms of accuracy metrics, MonoGAE significantly outperforms ego-vehicle monocular 3D object detection methods on the DAIR-V2X-I and Rope3D (homogeneous) datasets, achieving state-of-the-art results. In terms of robustness, our method also achieved SOTA results on the Rope3D dataset, demonstrating strong robustness and generalization performance in unknown road scenes. Our main contributions are summarized as follows: 

\begin{enumerate}
\item{we propose a road-side monocular 3D object detection method based on ground-aware embedding, which achieves higher detection accuracy and generalization performance by integrating implicit roadside ground information with high-dimensional semantic features.}
\item{In order to generate better implicit ground feature information, proposing a pixel-level ground plane equation map encoding method as the ground truth for the auxiliary branch Ground Predictor.}
\item{Conducting validation experiments on the DAIR-V2X and Rope3D datasets, our method significantly outperforms existing methods and achieves SOTA (state-of-the-art) results. Additionally, under the heterogeneous data partition of the Rope3D dataset, our method also outperformed existing methods, demonstrating strong robustness and generalization.}
\end{enumerate}
\section{Related Work}

\mypara{Monocular 3D object detection.} Monocular 3D object detection (Mono3D) seeks to anticipate 3D bounding boxes using an input image. The prevailing Mono3D techniques can be broadly categorized into three distinct groups. 1) Geometric Constraint-based Methods: This category encompasses approaches that leverage additional information regarding pre-existing 3D vehicle configurations. Widely employed resources include vehicle Computer-Aided Design (CAD) models \cite{liu2021autoshape,manhardt2019roi,Chabot2017manta} as well as key points \cite{barabanau2019monocular}. However, this approach necessitates incurring additional labeling costs. 2) Depth Assist Methods: This category involves the prediction of an independent depth map for the monocular image as the initial step. The depth map is then transformed into artificial dense point clouds so as to employ the existing 3D object detectors \cite{song2023graphalign++,song2023vp}. Such prior knowledge can be obtained through diverse avenues, including the generation of a depth map through LiDAR point cloud (or Pseudo-LiDAR) techniques \cite{wang2019pseudo,reading2021categorical}, utilization of monocular depth prediction models \cite{ma2020rethinking,ding2020learning}, or the generation of a disparity map via stereo cameras \cite{liu2021yolostereo3d}. However, the availability of such external data is not universally accessible across all scenarios. Moreover, the prediction of these dense heatmaps leads to a notable increase in inference time. 3) Pure Image-Based Methods: This category encompasses approaches that operate solely on the basis of the input image without the need for additional side-channel information. These techniques \cite{yang2023lite,fan2023calibration,zhou2019objects, yang2023mix} exclusively utilize a single image as input and embrace center-based pipelines that adhere to conventional 2D detectors \cite{zhou2019objects,tian2019fcos}. M3D-RPN \cite{brazil2019m3d} reconceptualizes the challenge of monocular 3D detection by presenting a dedicated 3D region proposal network. Notably, SMOKE \cite{liu2020smoke} and FCOS3D \cite{wang2021fcos3d} employ minimal handcrafted components to project a 3D bounding box prediction. They achieve this through a concise one-stage keypoint estimation procedure, coupled with the regression of 3D variables rooted in CenterNet \cite{zhou2019objects} and FCOS \cite{tian2019fcos}, respectively. In pursuit of enhancing the robustness of monocular detectors, leading-edge techniques have introduced more potent yet intricate geometric priors. MonoPair \cite{chen2020monopair} advances the modeling of occluded objects by accounting for the interplay among paired samples and interpreting spatial relations with a degree of uncertainty. Kinematic3D \cite{brazil2020kinematic} introduces an innovative methodology for monocular video-based 3D object detection, harnessing kinematic motion to refine the accuracy of 3D localization. MonoEF \cite{zhou2021monoef} introduces an inventive approach to capturing camera pose, enabling the formulation of detectors impervious to extrinsic perturbations. MonoFlex \cite{zhang2021objects} employs an uncertainty-guided depth ensemble strategy and categorizes distinct objects for tailored processing. MonoDLE \cite{ma2021delving} analyzes the bottlenecks of pure monocular detectors and designs dedicated components to address these issues. GUPNet \cite{lu2021geometry} tackles error amplification through geometry-guided depth uncertainty and employs a hierarchical learning strategy to mitigate training instability. MonoDETR \cite{zhang2022monodetr} presents a streamlined monocular object detection framework, endowing the conventional transformer architecture with depth awareness and mandating depth-guided supervision throughout the detection process. The aforementioned geometrically reliant designs significantly elevate the overall performance of center-based methods. However, it is important to note that the current methodologies predominantly concentrate on ego-vehicle autonomous driving scenarios, exhibiting a narrower emphasis on the utilization of Mono3D within roadside scenarios. Moreover, these methods tend to offer limited consideration to the challenges posed by the diverse camera orientations at different intersections, which can adversely impact the robustness of the Mono3D approach in such settings.

\mypara{Ground knowledge in monocular 3D object detection.} Several attempts have been made to utilize ground knowledge in monocular 3D object detection. Mono3D \cite{chen2016monocular} was the first to try using the ground plane to generate 3D bounding box proposals. GROUND-AWARE \cite{liu2021ground} introduced the ground plane in geometric mapping and proposed a ground-aware convolution module to enhance detection. MonoGround \cite{qin2022monoground} suggested replacing the bottom surface of the 3D bounding box with the ground plane, introducing depth information through ground plane priors, and proposing depth alignment training strategies and two-stage depth inference methods. MoGDE \cite{zhou2022mogde} envisioned a virtual 3D scene consisting of only the sky and the ground, where each pixel had associated depth information. This enabled MoGDE to utilize dynamic ground depth information as prior knowledge to guide Mono3D and improve detection accuracy. However, in these methods, the ground plane was defined based on a vehicle's viewpoint, and assumed all positions at a distance of 1.65 meters from the camera to be the ground plane \cite{chen2016monocular,liu2021ground}. Since the ground plane from a roadside viewpoint is not parallel to the camera's viewpoint, these methods are not applicable to roadside data. In this paper, we propose a refined ground plane equation map with camera extrinsic parameters and existing labels. Additionally, a ground feature module is introduced to produce high-dimensional ground-aware embeddings.
\section{Method}
\subsection{Problem Definition}
The focus of this work is to detect the three-dimensional bounding boxes of foreground objects within images. Specifically, given an image $I_{cam}\in \mathbb{R}^{H\times W\times 3}$ captured by roadside cameras, we can derive the extrinsic matrix $E \in \mathbb{R}^{3\times4}$, intrinsic matrix $K \in R^{3 \times 3}$ and ground plane equation $G \in \mathbb{R}^{4 \times 1}$ through camera calibration. Our objective is to precisely locate the 3D bounding boxes of the objects depicted in the image. These bounding boxes are collectively denoted as $B=\left\{B_1,B_2,…,B_n\right\}$, while the detector's output is represented as $\hat{B}$. Each individual 3D bounding box, labeled as $\hat{B}_{i}$, is defined as a seven-degree-of-freedom vector.	
\begin{equation}
    \hat{B}_{i} = \left(x, y, z, l, w, h, \theta \right),
    \label{con:eq1}
\end{equation}
where $\left(x,y,z \right)$ represents the coordinates of each 3D bounding box, and $\left(l,w,h \right)$ denotes the dimensions of the cuboid—length, width, and height, respectively. The variable $\theta$ indicates the yaw angle of each instance with respect to a designated axis. 

To provide a clearer definition, we can formulate a monocular 3D object detector, labeled as $F_{Mono3D}$, as follows:
\begin{equation}
    \hat{B} = F_{Mono3D}\left(I_{cam}\right).
    \label{con:eq2}
\end{equation}

\subsection{MonoGAE}
The core motivation of our method is utilizing the stable and strong ground plane prior knowledge to improve the performance of monocular 3D object detection in roadside scenes. There are two challenges: (1) Bridging the gap between ground geometry information and high-level image features, harmoniously fusing them. (2) designing a robust representation of the ground plane that remains effective despite the varying camera installation orientations across a range of roadside scenes.
To this end, we propose a straightforward framework for enhancing roadside monocular 3D object detection through the incorporation of ground-aware embeddings, dubbed MonoGAE.

\mypara{Overall Architecture.} As shown in \cref{fig_2}, our MonoGAE consists of an image backbone, a ground feature module, a visual feature module, a ground-guided decoder, and a detection head. The image backbone is responsible for extracting four 2D high-dimensional multi-scale feature maps $F=\left\{f_{1/8},f_{1/16},f_{1/32},f_{1/64}\right\}$ given an image $I_{cam}$. The visual feature module aims to generate the visual embeddings represented by $f_V^e \in \mathbb{R}^{S \times C}$, where $S$ is the sum of
the height and width of the four feature maps. Following a supervised training paradigm with the ground plane as labels, the ground feature module produces the high-level ground-aware ground embeddings denoted as $f_G^e \in \mathbb{R}^{\frac{HW}{16^2} \times C}$, where $H$, $W$ is the height and width of the input image, respectively. After obtaining the visual and ground embeddings, the ground-guided decoder combines these two embeddings together and generates enhanced object queries $Q_{GV} \in \mathbb{R}^{N\times C}$, where $N$ denotes the pre-defined maximum object number within an image. These queries will be further feed into the detection head to predict the 3D bounding box consisting of location $(x, y, z)$, dimension $(l, w, h)$, and orientation $\theta$. We will provide a detailed analysis of the representation method of the ground plane below.

\mypara{Visual Feature Module.} We combine four feature maps at various scales, each accompanied by sine/cosine positional encodings, resulting in a flattened image feature denoted as $f_V\in \mathbb{R}^{S\times C}$, where $S$ signifies the cumulative sum of the dimensions (height and width) of the four feature maps. This amalgamated feature is subsequently inputted into the Visual Encoder, leading to the generation of visual embeddings $f_V^e\in \mathbb{R}^{S\times C}$.

We apply three encoder blocks in the visual encoder, each block is composed of two main components: a self-attention layer and a feed-forward neural network (FFN). This configuration facilitates the capture of information spanning diverse spatial extents within the image, thereby enhancing both the expressive capacity and the distinctiveness of the visual information. We formulate the process of the the self-attention layer in visual block as,
\begin{equation}
f_V^{mid} = SelfAttn(f_V) = Concat(head_1, ..., head_h) W^O
\end{equation}
where $h$ is the number of multi head in self-attention layer,  $W^O \in \mathbb{R}^{C \times C}$ is the learnable weights of a linear layer.
\begin{eqnarray}
\label{eq}
head_i& =& Attention(Q_{f_V}, K_{f_V}, V_{f_V}) \nonumber   \\
~&=&Softmax\left(\frac{{Q_{f_V} {K_{f_V}}^T}}{{\sqrt{C}}}\right) V_{f_V}
\end{eqnarray} 
where $Q_{f_V}= f_V W_{f_{V}}^Q$, $K_{f_V} = f_V W_{f_{V}}^K$, $V_{f_V} = f_V W_{f_{V}}^V$, and then $W_{f_{V}}^Q \in \mathbb{R}^{C\times C_q}$, $W_{f_{V}}^K \in \mathbb{R}^{C\times C_k}$, $W_{f_{V}}^V \in \mathbb{R}^{C\times C_v}$, $C_q = C_k = C_v = C / h$. They are all learnable weights of projection layers.

The feed-forward neural network (FFN) consists of two linear transformations with a ReLU activation in between, which can be formulated as follows:
\begin{equation}
f_V^e = FFN(f_V^{mid})= Linear(ReLU(Linear(f_V^{mid}))) .
\end{equation}

\mypara{Ground Feature Module.}
Multi-scale features $f_{1/8}$, $f_{1/16}$, $f_{1/32}$ from the image backbone are unified to the same size feature maps with 1/16 resolution of the input image through nearest-neighbor sampling. All three feature maps are combined through element-wise addition, resulting in fused features possessing multi-scale information.	
Then, a convolutional layer is employed to extract the initial ground features denoted as $f_G\in \mathbb{R}^{\frac{H}{16}\times\frac{W}{16}\times C}$ from the fused features. 
Following this, the initial ground features $f_G$ are input to the ground encoder, resulting in the creation of ground-aware embeddings denoted as $f_G^e\in \mathbb{R}^{\frac{HW}{16^2}\times C}$. 
In the ground encoder, we utilize the same encoder block as employed in the visual encoder described above. The decoupling of the ground encoder and visual encoder allows them to enhance the learning of their distinct features, thereby enabling separate encoding of the visual and ground information for the input image. 
To enhance $f_G$ with more reliable ground plane information, we further input it to the ground predictor to predict the equation map of the ground plane, which we denote by $denorm^{\theta}$. The refined ground plane equation map that will be explained in detail below is used as the ground truth. The ground predictor is composed of two residual blocks as in ResNet \cite{he2016deep}. Considering the variation in camera positions across different intersections, a corrective ground plane equation map is introduced to address the challenges posed by this diversity. This map serves as the label for the ground predictor, enhancing robustness.

\mypara{Ground Plane Representation.} The ground plane equation is represented as $G_{initial}: \alpha X + \beta Y + \gamma Z + d = 0$, where $(\alpha, \beta, \gamma)$ represents the normal vector of the ground, and $d$ denotes the distance from the ground to the coordinate origin. Due to the fixed installation position of roadside cameras, the equation of the ground plane remains unchanged, and the existing datasets offer information regarding the ground plane equation. The subsequent sections primarily introduce three representations of the ground plane: ground plane depth map, ground plane equation map, and refined ground plane equation map.

\subsubsection{Ground depth map} By projecting the ground plane onto the image, we can generate a ground depth map in which the depth of each pixel is determined by both the camera's intrinsic parameters $K$ and the ground equation $G^{1\times 4}$. Given the pixel $(u, v)$ of the ground depth map, along with the ground equation $G^{1\times 4}$ and the camera's intrinsic parameters $K^{3\times 3}$, the 3D coordinates $(x, y, z)$ of the point within the camera coordinate system can be calculated using Eq. \ref{eq5}. Here, $z$ signifies the depth value of the specific point.

\begin{equation}
\begin{cases}  
&z \cdot \begin{bmatrix} u , v , 1 \end{bmatrix}^\mathsf{T} = K^{3\times3} \begin{bmatrix} x , y , z \end{bmatrix}^\mathsf{T}\\
&G^{1\times4} \begin{bmatrix} x , y , z , 1 \end{bmatrix}^\mathsf{T} = 0
\end{cases},
\label{eq5}
\end{equation}

\begin{figure}[!h]
\centering
\includegraphics[width=2.8in]{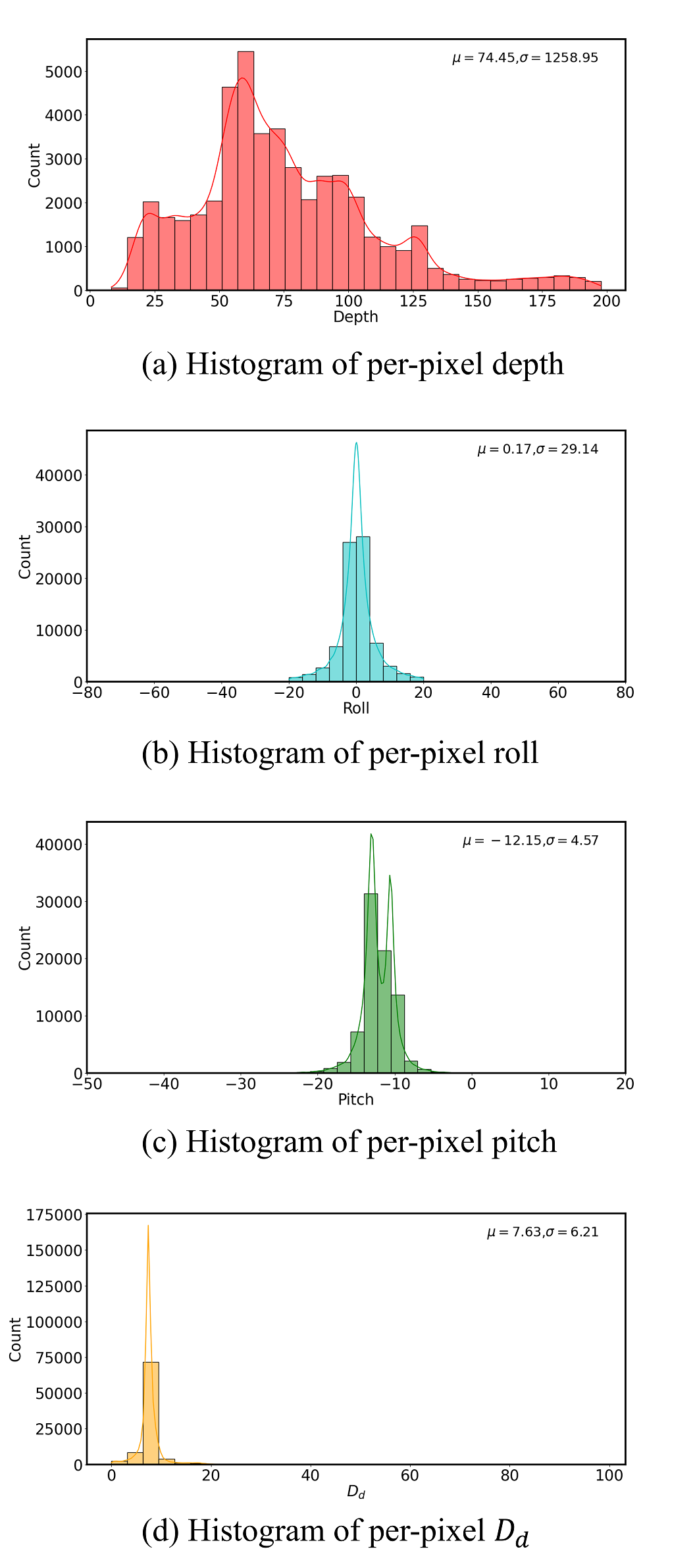}
\caption{\textbf{The comparison of predicting the depth map and the refined ground plane equation map.} (a) We plot the histogram of per-pixel depth. (b-d) We construct histograms illustrating per-pixel ground plane equations, which can alternatively be interpreted as the camera's mounting roll, pitch, and height.	It is evident that the depth range exceeds 200 meters, whereas the distribution of the camera's pose parameters is concentrated. This concentration simplifies the network to predict the refined ground plane equation.	}
\label{fig_3}
\end{figure}

Notably, the statistical analysis (refer to \cref{fig_3}~(a)) unveils a vehicle distance distribution within roadside scenarios spanning the range of $10m$ to $200m$. This range notably exceeds the scales observed in both the KITTI \cite{geiger2012we} and nuScenes \cite{Caesar2019nuScenesAM} datasets, considering the perspective of the ego-vehicle.

\begin{figure}[!h]
\centering
\includegraphics[width=3.5in]{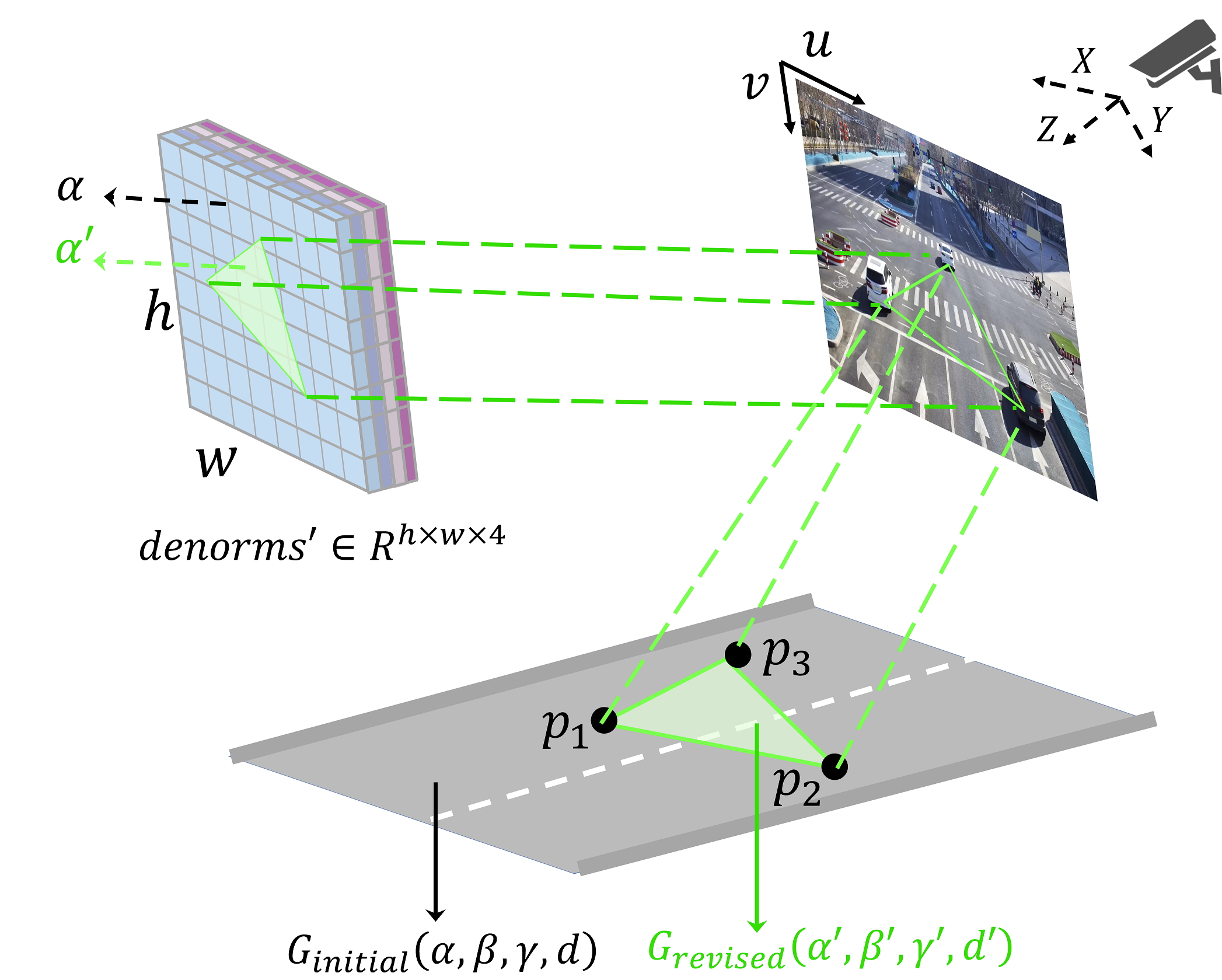}
\caption{\textbf{A diagram pipeline to get the corrected $denorms'$}. 
First, the global ground equation $G_{initial}$ is used to initialize the $denorms' \in \mathbb{R}^{h\times w \times 4}$, the four grids correspond to $\alpha$, $\beta$, $\gamma$, and $d$ from left to right.
Then, the sub ground planes are determined by the ground center point of 3D annotations, which can be used to further update the corresponding areas of $denorms'$.}
\label{fig_4}
\end{figure}
\subsubsection{Ground plane equation map} 
We divide the entire ground into multiple small grids, each grid has its corresponding set of four ground equation parameters: $\alpha$, $\beta$, $\gamma$, and $d$, and finally construct a pixel-level fine-grained ground plane equation map $denorms \in \mathbb{R}^{h\times w \times 4}$. Each pixel is assigned with the ground plane equation information corresponding to its associated ground grid.
Considering the complexity of the actual environment, real roads are not completely flat and without concave surfaces. Hence, initializing the ground plane equation map with the global plane equation is suboptimal, as it falls short of accurately simulating the complete real environment. In order to achieve a more precise representation of the ground, incorporating 3D annotations of vehicles becomes essential for implementing further refinements.

\subsubsection{Refined ground plane equation map} In practical scenarios, objects are typically situated on the ground plane, allowing us to approximate the center of an object's bottom surface as a point within the ground plane. Subsequently, we acquire a set of points denoted as $P$ from 3D annotations. In accordance with the theorem stating that ``three non-collinear points suffice to define a plane," we choose any three points, denoted as $p_1$, $p_2$, and $p_3$, from the set $P$. We then insert their spatial coordinates into Eq. \ref{eq6} to compute the revised ground equation $G_{revised}$: $\alpha' X + \beta' Y + \gamma' Z + d' = 0$.	Through the projection of these three points onto the image, we derive the corresponding pixel coordinates $(u_1, v_1)$, $(u_2, v_2)$, and $(u_3, v_3)$. Subsequently, leveraging these coordinates, we identify the triangular regions within the ground plane equation map that require refinement. Furthermore, we insert the four parameters of $G_{revised}$ into the respective region of $denorms' \in \mathbb{R}^{h\times w\times 4}$, as depicted in \cref{fig_4}. In order to minimize the discrepancy between the computed ground equation and the real-world environment, it is advisable to select three points with the smallest areas as the reference for computation. This approach can yield a ground plane equation that is more detailed and less prone to errors.
\begin{equation}
\label{eq6}
\begin{cases}
&\alpha' = (y_2-y_1 )(z_3-z_1 )-(y_3-y_1 )(z_2-z_1 )\\
&\beta' = (z_2-z_1 )(x_3-x_1 )-(z_3-z_1 )(x_2-x_1 ) \\
&\gamma' = (x_2-x_1 )(y_3-y_1 )-(x_3-x_1 )(y_2-y_1 )\\
&d' = -\alpha' \cdot x_1-\beta' \cdot y_1-\gamma' \cdot z_1 
\end{cases},
\end{equation}

\begin{figure*}[h!t]
\centering
\includegraphics[width=0.99\textwidth]{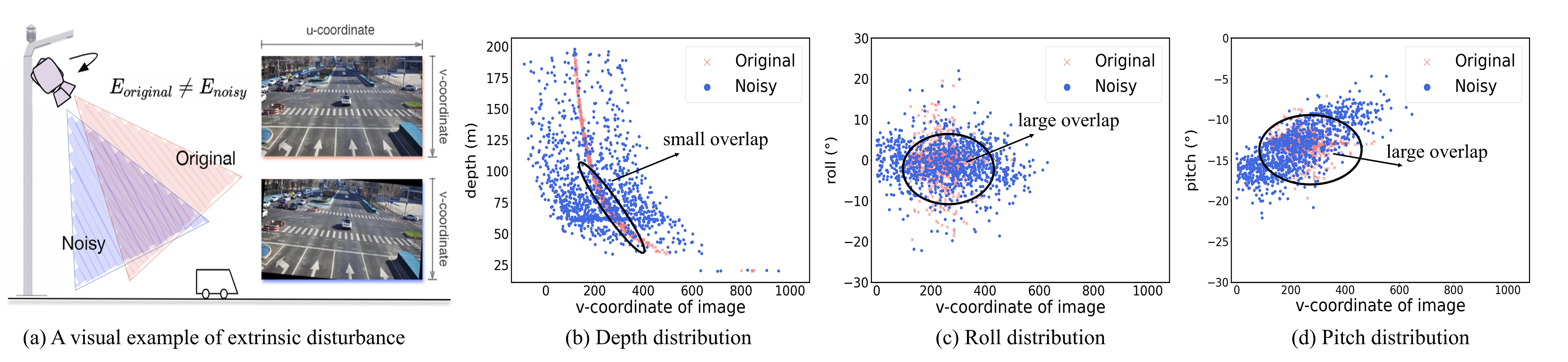}
\caption{\textbf{The correlation between the object's row coordinates on the image with its depth, roll, and pitch.} The position of the object in the image, which can be defined as (u, v), and the v-coordinate denotes its row coordinate of the image. (a) A visual example of the noisy setting, adding a rotation offset along roll and pitch directions in the normal distribution. (b) is the scatter diagram of the depth distribution. (c) is for the roll from the ground.(d) is for the pitch from the ground. We can find, compared with depth, the noisy setting of roll and pitch have larger overlap with its original distribution, which demonstrates height estimation is more robust.}
\label{fig_5}
\end{figure*}

\subsubsection{Comparing the depth map and refined equation map of ground plane}
we leverage the 3D annotations of the DAIR-V2X-I \cite{yu2022dair} dataset, where we first project the bottom center point of 3D bounding boxes to the images, to plot the histogram of per-pixel depth in \cref{fig_3}~(a). We can observe a large range from 0 to 200 meters. By contrast, we conducted a similar histogram analysis on the refined ground plane equation map. To facilitate comparative analysis, the ground plane equation in the camera coordinate system is converted into the roll, pitch, yaw, and distance $D_d$ of the camera relative to the ground plane. The histogram for roll, pitch, and distance are shown in \cref{fig_3} (b-d), revealing noticeably smaller intervals, which is easier for the network to predict.

\cref{fig_5} (a) offers a visual illustration of extrinsic disturbance. This visual example serves to demonstrate the superiority of predicting the ground plane equation over ground depth. To show that predicting the refined ground plane equation map is superior to the depth map, we plot the scatter graph to show the correlation between the object's row coordinates on the image and its depth in \cref{fig_5} (b). Each point represents an instance. Consistently with the previous, the equation of the ground plane in the camera coordinate system is transformed into the camera's roll and pitch relative to the ground plane. We also plot the scatter graph to show the correlation between the object's row coordinates on the image and its roll and pitch relative to the ground plane in \cref{fig_5} (c-d). As shown in \cref{fig_5} (b), we observe a clear trend: objects with smaller depths exhibit smaller v values. However, when the extrinsic parameters undergo variation, a comparison between the same metric plotted in blue reveals significantly divergent values from the pristine configuration. In this scenario, where only minimal overlap exists between the clean and noisy configurations, it becomes evident that predicting a ground depth map would result in performance deterioration with changing external parameters. Conversely, as evidenced in \cref{fig_5} (c-d), the distribution remains relatively consistent irrespective of alterations in external parameters; specifically, the overlap between the orange and blue data points is substantial. This observation compels us to consider utilizing the equation map rather than the depth map to represent the ground plane. By adopting this approach, our method effectively maintains strong robustness against the wide-ranging camera roll and pitch angles encountered at various intersections.

\mypara{Ground-guided Decoder.}
The module serves the purpose of effectively fusing visual embeddings $f_{V}^{e}$ and ground embeddings $f_{G}^{e}$. 
We apply three ground-guided decoder blocks, each of which consists of a ground cross-attention layer, a self-attention layer, a visual cross-attention layer, and a feedforward neural network (FFN).
We employ a learnable object query $q \in \mathbb{R}^{N\times C}$ to adaptively capture geometric cues from the ground embeddings and semantic features from visual embeddings. 

The ground cross-attention layer empowers each query $q$ to dynamically explore geometric cues within the ground region of the image. This capability aids in gaining a more comprehensive understanding of scene-level spatial information and facilitates the modeling of geometric relationships among objects. The specific process of producing the ground-aware object queries $Q_G \in \mathbb{R}^{N\times C}$ through the ground cross-attention layer can be formulated as follows.
\begin{eqnarray}
\label{ground_cross_attention}
Q_G& = &CrossAttn(q, f_G^e) \nonumber   \\
&=& Concat(head_1^{'}, ..., head_h^{'}) W^O,
\end{eqnarray}
where $h$ is the number of multi head in the ground cross attention layer,  $W^O \in \mathbb{R}^{C \times C}$ is the learnable weights of a linear layer.
\begin{eqnarray}
\label{eq}
head_i^{'}& =& Attention(Q_q, K_G, V_G) \nonumber   \\
~&=&Softmax\left(\frac{{Q_q \left(K_G\right)^T}}{{\sqrt{C}}}\right) V_G \nonumber   \\
~&=&A_G V_G
\end{eqnarray} 
where $Q_q=Linear(q) \in \mathbb{R}^{N\times C}$, $K_{G}$ is obtained by $K_{G}=Linear(f_G^e) \in \mathbb{R}^{\frac {HW}{16^2} \times C}$, 
and $V_{G}$ is obtained through $V_{G} = Linear(f_G^e) \in \mathbb{R}^{\frac {HW}{16^2} \times C}$, $A_{G} \in \mathbb{R}^{N\times \frac {HW}{16^2}}$ is the query-ground attention map.

Subsequently, $Q_G$ is inputted into a self-attention layer and for further interaction, avoiding redundant predictions of the same object's bounding boxes. This process can be formulated as follows:
\begin{equation}
\label{self_attention}
Q_G = SelfAttn(Q_G),
\end{equation}

Finally, the visual cross-attention layer alongside an additional FFN layer further enhances the visual feature embeddings $f_V^e$ for object queries, together with a FFN layer, resulting in augmented object queries denoted as $Q_{GV} \in \mathbb{R}^{N \times C}$.
\begin{equation}
\label{visual_cross_attention}
Q_{GV}^{mid} = CrossAttn(Q_G, f_V^e),
\end{equation}
\begin{eqnarray}
\label{ffn}
Q_{GV}&=&FFN(Q_{GV}^{mid}) \nonumber   \\
~&=&Linear(ReLU(Linear(Q_{GV}^{mid}))).
\end{eqnarray}

Through this ground-guided decoding process, two kinds of embedded information features can be seamlessly integrated, resulting in a substantial enhancement of the 3D attribute prediction performance for each object query. This improvement transcends the previous limitations imposed by the finite visual features around the center.

\mypara{Training Loss.}
MonoGAE is an end-to-end network in which all components are jointly trained based on a composite loss function comprising $L_{2D}$, $L_{3D}$, and $L_{denorm}$. Specifically, the 2D object loss $L_{2D}$ primarily concerns the 2D visual appearance of images, using Focal loss \cite{lin2017focal} to estimate the object classes, L1 loss to estimate the 2D size $(l,r,t,b)$ and projection of the 3D center $(x_{3d},y_{3d})$, and GIoU loss for 2D box IoU. Finally, $L_{2D}$ can be represented as:
\begin{equation}
L_{2D} = \omega_1 L_{class} + \omega_2 L_{2dsize} + \omega_3 L_{xy3d} + \omega_4 L_{giou},
\end{equation}

The main focus of $L_{3D}$ is on the 3D spatial properties of objects. L1 loss is utilized to estimate the 3D dimensions $(h_{3d},w_{3d},l_{3d})$ as well as the orientation angle. For the depth value $d_{pre}$, the final depth loss is formed by using the Laplace arbitrary uncertainty loss\cite{chen2020monopair}:
\begin{equation}
L_{depth} = \frac{2}{\sigma} |d_{gt} - d_{pre}| + \log(\sigma),
\end{equation}
where $\sigma$ is the standard deviation predicted together with $d_{pre}$, and $d_{gt}$ is the actual depth value of the ground truth. Overall, $L_{3D}$ can be expressed as:
\begin{equation}
L_{3D} = \omega_5 L_{3dsize} + \omega_6 L_{angle} + \omega_7 L_{depth},
\end{equation}

The loss function $L_{denorm}$ between the ground plane equation map $denorms^\theta$ predicted based on $f_G$ and the refined ground plane equation map $denorms'$ is:
\begin{equation}
L_{denorm} = \frac{1}{h \times w \times 4} \sum_{}{} \left| denorms^{\theta} - denorms' \right|,
\end{equation}
The overall loss formula is: 
\begin{equation}
L = L_{2D} + L_{3D} + \omega_8 L_{denorm}.
\end{equation}
where $\omega_1$ to $\omega_8$ are balancing weights.
\section{Experiments}
\subsection{Settings}
\mypara{Dataset.} 
We perform experiments on two roadside datasets: DAIR-V2X \cite{yu2022dair} and Rope3D \cite{ye2022rope3d}. The DAIR-V2X dataset encompasses images captured from both vehicles and roadside units. Here, we focus on the DAIR-V2X-I, a subset exclusively composed of images obtained from mounted cameras, thereby centering our study on roadside perception. Specifically, the DAIR-V2X-I dataset encompasses approximately 10,000 images, with 50\% allocated for training, 20\% for validation, and 30\% for testing purposes. we mainly used the 3D average precision AP$_{3D}$$|\scriptstyle R40$  \cite{simonelli2019disentangling} as the evaluation metric, analogous to the approach employed in the KITTI \cite{geiger2012we} dataset.
Rope3D \cite{ye2022rope3d} is another extensive dataset, encompassing more than 500,000 images collected from a total of seventeen intersections. In line with the suggested homologous configuration, we allocate 70\% of the images for training, reserving the remainder for validation. To assess performance, we employ the same AP$_{3D}$$|\scriptstyle R40$ as in \cite{geiger2012we} and the $Rope_{score}$ as depicted in \cite{ye2022rope3d}, which is a composite metric derived from AP$_{3D}$$|\scriptstyle R40$ and other similarity metrics, including average ground center similarity, average orientation similarity, average area similarity and average four ground points distance and similarity. 

\mypara{Training Details.} 
We employ ResNet-50 \cite{he2016deep} as the image backbone, with an input image resolution of $512 \times 928$. Random horizontal flip data augmentation is applied. The number of object queries $q$, is set to 100. The balance weights $\omega_1$ to $\omega_8$ in the training loss are configured as follows: 2, 10, 5, 2, 1, 1, 1, and 1. The AdamW optimizer is utilized with a learning rate of $2 \times 10^{-4}$ and a weight decay of $1 \times 10^{-5}$. The batch size is set to 8, and the training epoch is fixed at 200. The learning rate is decreased by a factor of 0.1 at the 125th and 160th epochs.

\begin{table*}[!t]
\small
\centering
\renewcommand{\arraystretch}{1.2}
\setlength\tabcolsep{10.8pt}
\caption{\textbf{Comparing with the state-of-the-art on the DAIR-V2X-I val set.} Here, we report the results of three types of objects, vehicle~(veh.), pedestrian~(ped.) and cyclist~(cyc.). Each object is categorized into three settings according to the difficulty defined in ~\cite{yu2022dair}. $\dagger$ indicates methods specifically designed for monocular 3D object detection. $\ast$ signifies frameworks tailored for multi-view 3D object detection.}
\label{tab:table1}
\begin{tabular}{l|c|ccc|ccc|ccc}
\hline
\multirow{2}{*}{Method} & \multirow{2}{*}{Modal} & \multicolumn{3}{c|}{$Veh._{(IoU=0.5)}$} & \multicolumn{3}{c|}{$Ped._{(IoU=0.25)}$} & \multicolumn{3}{c}{$Cyc._{(IoU=0.25)}$} \\ \cline{3-11} 
                        &                    & Easy       & Mod.      & Hard      & Easy       & Mod.      & Hard      & Easy      & Mod.      & Hard      \\ \hline
PointPillars\cite{lang2019pointpillars}            & L                  & 63.07      & 54.00     & 54.01     & 38.54      & 37.21     & 37.28     & 38.46     & 22.60     & 22.49     \\
SECOND\cite{yan2018second}                  & L                  & 71.47      & 53.99     & 54.00     & 55.17      & 52.49     & 52.52     & 54.68     & 31.05     & 31.20     \\
MVXNET(PF)\cite{sindagi2019mvx}              & C+L                & 71.04      & 53.72     & 53.76     & \textbf{55.83}      & \textbf{54.46}     & \textbf{54.40}     & 54.05     & 30.79     & 31.07     \\ \hline
Imvoxelnet\cite{rukhovich2022imvoxelnet}$\ast$             & C                  & 44.78      & 37.58     & 37.56     & 6.81       & 6.75      & 6.73      & 21.06     & 13.58     & 13.18     \\
MonoDETR\cite{zhang2022monodetr}$\dagger$          & C                  & 58.86      & 51.24     & 51.00     & 16.30       & 15.37      & 15.61      & 37.93     & 34.04     & 33.98     \\
BEVFormer\cite{li2022bevformer}$\ast$              & C                  & 61.37      & 50.73     & 50.73     & 16.89      & 15.82     & 15.95     & 22.16     & 22.13     & 22.06     \\
BEVDepth\cite{li2022bevdepth}$\ast$               & C                  & 75.50      & 63.58     & 63.67     & 34.95      & 33.42     & 33.27     & 55.67     & 55.47     & 55.34     \\
BEVHeight\cite{yang2023bevheight}$\ast$
               & C                  & 77.78      & 65.77     & 65.85     & 41.22      & 39.39     & 39.46     & 60.23     & 60.08     & 60.54 \\
BEVHeight++\cite{yang2023bevheight_plus}$\ast$
               & C                  & 79.31      & 68.62     & 68.68     & 42.87      & 40.88     & 41.06     & \textbf{60.76}     & \textbf{60.52}     & \textbf{61.01}     \\ \hline
\rowcolor{cyan!30} Ours$\dagger$                    & C                  & \textbf{84.61}      & \textbf{75.93}     & \textbf{74.17}     & 25.65      & 24.28     & 24.44     & 44.04     & 47.62     & 46.75     \\ \hline
\end{tabular}
\end{table*}
\subsection{Comparing with state-of-the-art}

\mypara{DAIR-V2X benchmark.} 
On the DAIR-V2X-I benchmark, we compare our method with other state-of-the-art approaches, namely MonoDETR \cite{zhang2022monodetr}, ImvoxelNet \cite{rukhovich2022imvoxelnet}, BEVFormer \cite{li2022bevformer}, and BEVDepth \cite{li2022bevdepth}. Additionally, we present certain outcomes obtained from LiDAR-based and multimodal methods, as reproduced by the original DAIR-V2X \cite{yu2022dair} benchmark. The results can be seen from Tab.~\ref{tab:table1}. For the vehicle category, which encompasses car, truck, van, and bus, our proposed MonoGAE outperforms state-of-the-art BEVHeight++\cite{yang2023bevheight_plus} by substantial margins of 5.3\%, 7.31\%, and 5.49\% in the `Easy', `Mod', and `Hard' settings, respectively. When considering the pedestrian and cyclist categories, the challenges are amplified due to their smaller sizes and non-rigid body nature. However, our method still surpasses the MonoDETR\cite{zhang2022monodetr} baseline by 8.46\% and 25.48\%, respectively. These improvements demonstrate that strong prior information on the ground plane can significantly enhance the accuracy of monocular 3D object detection.

\begin{table}[!t]
\centering
\small
\renewcommand{\arraystretch}{1.2}
\setlength\tabcolsep{9.0pt}
\caption{\textbf{Results on the Rope3D val set based on homologous partition.} Here, we follow~\cite{ye2022rope3d} to report the results on vehicles. $\dagger$ indicates methods specifically designed for monocular 3D object detection. $\ast$ signifies frameworks tailored for multi-view 3D object detection.}
\label{tab:table2}
\begin{tabular}{l|cc|cc}
\hline
\multirow{2}{*}{Method} & \multicolumn{2}{c|}{Car} & \multicolumn{2}{c}{Big Vehicle} \\ \cline{2-5} 
                        & AP         & Rope       & AP             & Rope           \\ \hline
M3D-RPN\cite{brazil2019m3d}$\dagger$                 & 54.19      & 62.65      & 33.05          & 44.94          \\
Kinematic3D\cite{brazil2020kinematic}$\dagger$             & 50.57      & 58.86      & 37.60          & 48.08          \\
MonoDLE\cite{ma2021delving}$\dagger$                 & 51.70      & 60.36      & 40.34          & 50.07          \\
MonoFlex\cite{zhang2021objects}$\dagger$                & 60.33      & 66.86      & 37.33          & 47.96          \\
BEVFormer\cite{li2022bevformer}$\ast$               & 50.62      & 58.78      & 34.58          & 45.16          \\
BEVDepth\cite{li2022bevdepth}$\ast$                & 69.63      & 74.70      & 45.02          & 54.64          \\
BEVHeight\cite{yang2023bevheight}$\ast$               & 74.60      & 78.72      & 48.93          & 57.70          \\
BEVHeight++\cite{yang2023bevheight_plus}$\ast$               & 76.12      & 80.91      & 50.11          & 59.92          \\ \hline
\rowcolor{cyan!30}Ours$\dagger$                    & \textbf{80.12}&    \textbf{83.76}&      \textbf{54.62} &  \textbf{62.37}      \\ 
\bottomrule
\multicolumn{5}{l}{\footnotesize{AP and Rope denote AP$_{\text{3D}{|\text{R40}}}(IoU=0.5)$ and Rope$_\text{score}$ respectively.}}
\end{tabular}
\end{table}

\begin{figure*}[h!t]
	\centering
	\includegraphics[width=1\textwidth]{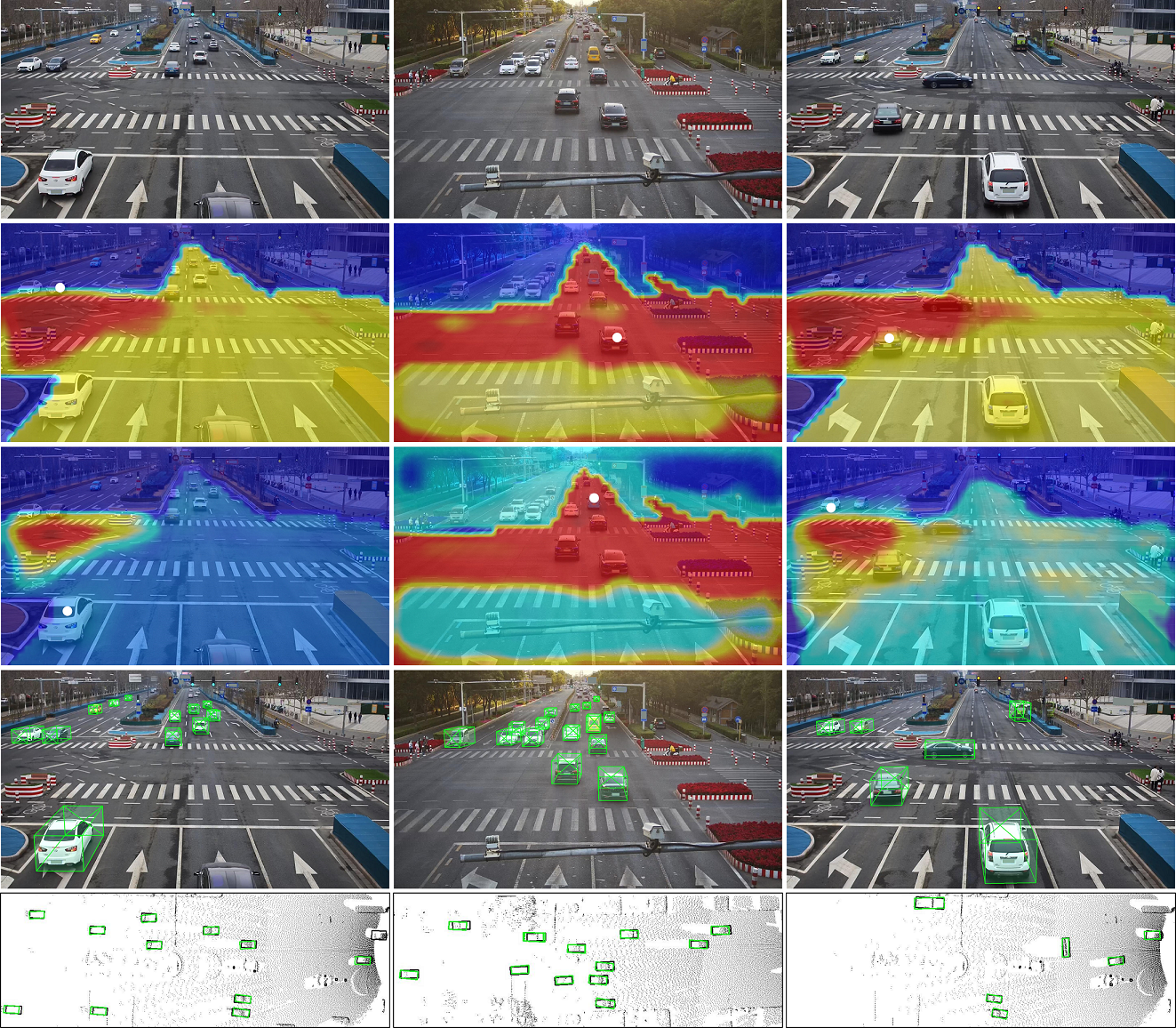}
	\caption{
	\textbf{Visualization of predictions and the attention maps $A_G$ in the ground cross-attention layer} The top row represent the input images and the bottom two rows represent the results of object detection. The middle two rows display attention maps corresponding to the target queries, indicated by white dots. Warmer colors represent stronger attention weights.}
\label{fig_6}
\end{figure*}
\mypara{Rope3D benchmark.} 
When evaluated on the Rope3D dataset, we conduct comparisons of our MonoGAE with other prominent methods, including MonoFlex\cite{zhang2021objects}, BEVFormer\cite{li2022bevformer}, BEVDepth\cite{li2022bevdepth}, BEVHeight\cite{yang2023bevheight} and BEVHeight++\cite{yang2023bevheight_plus} The results, as depicted in Table~\ref{tab:table2}, illustrate our method's superiority over all monocular and multi-view 3D object detectors listed in the table.

\subsection{Ablation Study}
We reported the $AP_{(3D|R40)}$ results of the "Vehicle" category on the DAIR-V2X-I validation set for all ablation studies. These results were achieved by modifying various components of the final solution.

\begin{table}[!t]
\centering
\small
\renewcommand{\arraystretch}{1.2}
\setlength\tabcolsep{12.5pt}
\caption{\textbf{Ablation study on the Ground Feature Module.} `GP' represents the ground predictor, `CL' implies convolution layers, and `GE' denotes the ground encoder.
}
\label{tab:table4}
\begin{tabular}{ccc|ccc}
\hline
GP  & CL&  GE& Easy  & Mod.  & Hard  \\ 
\hline
~ & ~& ~&         79.12 & 66.36 & 66.35 \\
$\checkmark$ &  & ~&          84.62& 71.58& 71.60 \\
$\checkmark$ & $\checkmark$& ~&         80.46 & 72.12 & 71.98 \\
$\checkmark$ & $\checkmark$& $\checkmark$& \textbf{84.61} & \textbf{75.93} & \textbf{74.17} \\ \hline
\end{tabular}
\end{table}
\mypara{Ground Feature Module.} We conduct ablation experiments on the configuration of the ground feature module. As shown in Tab.~\ref{tab:table4}, we test each component independently and report its performance. The overall baseline starts from 66.36\% $AP_{3D}$ on the moderate level. When the ground predictor is applied, the average precision is raised by 5.22\%points, Then, we add the convolution layers, which brings us a 0.54\% $AP_{3D}$ enhancement. Finally, the $AP_{3D}$ achieves 75.93\% when all components are applied, yielding a 3.81\% absolute improvement, validating the effectiveness of the ground feature module.

\begin{table}[!t]
\centering
\small
\renewcommand{\arraystretch}{1.2}
\setlength\tabcolsep{5.0pt}
\caption{\textbf{Ablation study on the ground plane representation.}\label{tab:table5}}
\begin{tabular}{l|lll}
\hline
Settings & Easy  & Mod.  & Hard  \\ \hline
(a) ground depth map      & 78.69 & 67.78 & 67.70 \\
(b) ground plane equation map     & 82.00 & 73.76 & 73.75 \\
(c) refined ground plane equation map     & \textbf{84.61} & \textbf{75.93} & \textbf{74.17} \\ \hline
\end{tabular}
\end{table}
\begin{table}[!t]
\centering
\small
\renewcommand{\arraystretch}{1.2}
\setlength\tabcolsep{5.0pt}
\caption{\textbf{Ablation study on the robustness against to various camera installation poses} ``roll” and ``pitch” means applying an additional rotation offset in normal distribution N(0, 0.3) to the camera’s extrinsic matrix along roll and pitch directions.}
\label{tab:disturb}
\begin{tabular}{l|cc|ccc}
\hline
Settings   &  roll  & pitch & Easy  & Mod.  & Hard  \\ \hline
\multirow{3}{*}{(a) ground depth map} & $\checkmark$  &    & 53.98 & 46.96 & 46.93 \\
&   & $\checkmark$   & 56.72 & 49.37 & 49.26 \\
&$\checkmark$    & $\checkmark$  & 48.47 & 41.85 & 41.84 \\
\hline
\multirow{3}{*}{(b) ground plane equation} & $\checkmark$  &    & 62.28 & 55.13 &  54.99 \\
&   & $\checkmark$   & 67.79 & 58.44 & 58.41 \\
\quad \; map &$\checkmark$    & $\checkmark$  & 53.89 & 49.94 & 49.91 \\
\hline
\multirow{3}{*}{(c) refined ground plane} & $\checkmark$  &    & 64.50 & 56.94 & 56.92 \\
&   & $\checkmark$   & 70.57 & 62.35 & 62.27 \\
\quad \ equation map &$\checkmark$    & $\checkmark$  & 59.80 & 51.02 & 51.01 \\
\hline
\end{tabular}
\end{table}

\begin{table}[!t]
\centering
\small
\renewcommand{\arraystretch}{1.2}
\setlength\tabcolsep{8.5pt}
\caption{\textbf{Ablation study on the number of encoder or decoder blocks in each module} `GE' denotes the ground encoder, `VE' represents the visual encoder and `GD' implies the ground-guided decoder.}
\label{tab:table6}
\begin{tabular}{l|c|ccc}
\hline
  Blocks                                    & Set. & Easy  & Mod.  & Hard  \\ \hline
\multirow{3}{*}{Encoder Blocks in VE} & 2    & 78.76 & 69.88 & 68.31 \\
                                       & 3    & \textbf{84.61} & \textbf{75.93} & \textbf{74.17} \\
                                       & 4    & 82.47 & 74.07 & 73.96 \\ \hline
\multirow{3}{*}{Encoder Blocks in GE} & 1    & \textbf{84.61} & \textbf{75.93} & \textbf{74.17}\\
                                       & 2    & 82.53  & 74.39 & 74.29 \\
                                       & 3    & 82.23 & 74.22 & 74.18 \\ \hline

\multirow{3}{*}{Decoder Blocks in GD}   & 2    & 78.99 & 69.90 & 68.20 \\
                                       & 3    & \textbf{84.61} & \textbf{75.93} & \textbf{74.17} \\
                                       & 4    & 82.77 & 74.05 & 72.29 \\ \hline
\end{tabular}
\end{table}

\mypara{Ground Plane Representations.} 
As shown in Tab.~\ref{tab:table5}, we conducted ablation experiments on the ground plane representation. We employed the following encoding methods: (a) ground plane depth map, (b) ground plane equation map initialized with the global plane equation, and (c) refined ground plane equation map further improved through 3D annotations for each image. Our observations reveal that utilizing the refined ground plane equation map produced the best results, indicating the superiority of our proposed refined ground plane equation map.

\mypara{Robustness to various camera installation poses.}
In real-world scenarios, camera parameters undergo frequent changes due to various factors. In this way, we ablate the robustness of ground plane representations (depth map and refined equation map) separately in such dynamically changing environments. We follow the approach outlined in \cite{yu2022benchmarking} to simulate scenarios involving alterations in external parameters. Specifically, we introduce a random rotational offset drawn from a normal distribution N(0, 0.3) along the roll and pitch axes. This is done considering that mounting points typically remain consistent.
During the evaluation process, we incorporate the introduced rotational offsets along the roll and pitch directions into the original extrinsic matrix. Subsequently, we apply rotation and translation operations to the image to uphold the calibration relationship between the new external reference and the image. The results, as demonstrated in Tab.~\ref{tab:disturb}, under the disturbance of roll and pitch, the ground plane equation map outperforms the ground depth map by 8.09\%. Moreover, the refined ground plane equation map exhibits a significant advantage over the ground depth map by 9.17\%, underscoring its robustness in scenarios with external camera perturbations.

\mypara{The Number of Encoder or Decoder Blocks.} we ablate the configuration of the visual encoder, the ground encoder, and the ground-guided encoder. As shown in Tab.~\ref{tab:table6}, it can be seen that MonoGAE achieved the best performance by using three encoder blocks in the visual encoder, one encoder block in the ground encoder, and three decoder blocks in the ground-guided decoder.

\subsection{Visualization Results}
To facilitate comprehension of our ground-aware framework, we visualize the attention maps of the ground cross-attention within the ground-guided decoder. In Fig.~\ref{fig_6}, we highlight the query points by coloring them in white. As depicted, the region of interest for each query extends across the entire expanse of the road areas. Since all objects are situated on the road, there exhibit a strong correlation between road features and the distance of these objects. This observation signifies that object queries can leverage ground information within our ground-guided pipeline, thereby enhancing their predictive capacity and overcoming the prior constraint imposed by restricted neighboring features around the center.

\section{Conclusion}
In this paper, we propose MonoGAE, a robust framework for roadside monocular 3D object detection with ground-aware embeddings, which can effectively utilize the ground plane prior knowledge in roadside scenarios to improve the performance of monocular 3D object detection. In particular, we employ a supervised training paradigm that utilizes the ground plane as labels, aiming to narrow the domain gap between ground geometry information and high-dimensional image features. Furthermore, we introduce a refined ground plane equation map as the representation of the ground plane, enhancing the detector's robustness to variations in cameras' installation poses. Through extensive experimentation concerning vehicle instances, our method surpasses all state-of-the-art approaches and achieves the highest performance, securing the top position in both DAIR-V2X-I and Rope3D benchmarks. We aspire for our work to illuminate the exploration of more effective utilization of the substantial prior information present in roadside scenes.

\section*{Acknowledgments}
This work was supported by the National High Technology Research and Development Program of China under Grant No. 2018YFE0204300, and the
National Natural Science Foundation of China under Grant No. 62273198, U1964203 and 52221005.

{\small
\bibliographystyle{ieee_fullname}
\bibliography{egbib}

\begin{thebibliography}{10}\itemsep=-1pt

\bibitem{9228884}
Eduardo Arnold, Mehrdad Dianati, Robert de Temple, and Saber Fallah.
\newblock Cooperative perception for 3d object detection in driving scenarios using infrastructure sensors.
\newblock {\em IEEE Transactions on Intelligent Transportation Systems}, 23(3):1852--1864, 2022.

\bibitem{barabanau2019monocular}
Ivan Barabanau, Alexey Artemov, Evgeny Burnaev, and Vyacheslav Murashkin.
\newblock Monocular 3d object detection via geometric reasoning on keypoints.
\newblock {\em arXiv preprint arXiv:1905.05618}, 2019.

\bibitem{brazil2019m3d}
Garrick Brazil and Xiaoming Liu.
\newblock M3d-rpn: Monocular 3d region proposal network for object detection.
\newblock In {\em Proceedings of the IEEE/CVF International Conference on Computer Vision}, pages 9287--9296, 2019.

\bibitem{brazil2020kinematic}
Garrick Brazil, Gerard Pons-Moll, Xiaoming Liu, and Bernt Schiele.
\newblock Kinematic 3d object detection in monocular video.
\newblock In {\em Computer Vision--ECCV 2020: 16th European Conference, Glasgow, UK, August 23--28, 2020, Proceedings, Part XXIII 16}, pages 135--152. Springer, 2020.

\bibitem{Caesar2019nuScenesAM}
Holger Caesar, Varun Bankiti, Alex~H. Lang, Sourabh Vora, Venice~Erin Liong, Qiang Xu, Anush Krishnan, Yu Pan, Giancarlo Baldan, and Oscar Beijbom.
\newblock nuscenes: A multimodal dataset for autonomous driving.
\newblock {\em 2020 IEEE/CVF Conference on Computer Vision and Pattern Recognition (CVPR)}, pages 11618--11628, 2019.

\bibitem{Chabot2017manta}
Florian Chabot, Mohamed Chaouch, Jaonary Rabarisoa, Celine Teuliere, and Thierry Chateau.
\newblock Deep manta: A coarse-to-fine many-task network for joint 2d and 3d vehicle analysis from monocular image.
\newblock In {\em 2017 IEEE Conference on Computer Vision and Pattern Recognition (CVPR)}, Jul 2017.

\bibitem{10061347}
Wei Chen, Jie Zhao, Wan-Lei Zhao, and Song-Yuan Wu.
\newblock Shape-aware monocular 3d object detection.
\newblock {\em IEEE Transactions on Intelligent Transportation Systems}, 24(6):6416--6424, 2023.

\bibitem{chen2016monocular}
Xiaozhi Chen, Kaustav Kundu, Ziyu Zhang, Huimin Ma, Sanja Fidler, and Raquel Urtasun.
\newblock Monocular 3d object detection for autonomous driving.
\newblock In {\em Proceedings of the IEEE conference on computer vision and pattern recognition}, pages 2147--2156, 2016.

\bibitem{chen2020monopair}
Yongjian Chen, Lei Tai, Kai Sun, and Mingyang Li.
\newblock Monopair: Monocular 3d object detection using pairwise spatial relationships.
\newblock In {\em Proceedings of the IEEE/CVF Conference on Computer Vision and Pattern Recognition}, pages 12093--12102, 2020.

\bibitem{ding2020learning}
Mingyu Ding, Yuqi Huo, Hongwei Yi, Zhe Wang, Jianping Shi, Zhiwu Lu, and Ping Luo.
\newblock Learning depth-guided convolutions for monocular 3d object detection.
\newblock In {\em Proceedings of the IEEE/CVF Conference on computer vision and pattern recognition workshops}, pages 1000--1001, 2020.

\bibitem{fan2023calibration}
Siqi Fan, Zhe Wang, Xiaoliang Huo, Yan Wang, and Jingjing Liu.
\newblock Calibration-free bev representation for infrastructure perception.
\newblock {\em arXiv preprint arXiv:2303.03583}, 2023.

\bibitem{fan2023quest}
Siqi Fan, Haibao Yu, Wenxian Yang, Jirui Yuan, and Zaiqing Nie.
\newblock Quest: Query stream for vehicle-infrastructure cooperative perception.
\newblock {\em arXiv preprint arXiv:2308.01804}, 2023.

\bibitem{geiger2012we}
Andreas Geiger, Philip Lenz, and Raquel Urtasun.
\newblock Are we ready for autonomous driving? the kitti vision benchmark suite.
\newblock In {\em 2012 IEEE conference on computer vision and pattern recognition}, pages 3354--3361. IEEE, 2012.

\bibitem{9780191}
Muhamad~Amirul Haq, Shanq-Jang Ruan, Mei-En Shao, Qazi Mazhar~Ul Haq, Pei-Jung Liang, and De-Qin Gao.
\newblock One stage monocular 3d object detection utilizing discrete depth and orientation representation.
\newblock {\em IEEE Transactions on Intelligent Transportation Systems}, 23(11):21630--21640, 2022.

\bibitem{he2016deep}
Kaiming He, Xiangyu Zhang, Shaoqing Ren, and Jian Sun.
\newblock Deep residual learning for image recognition.
\newblock In {\em Proceedings of the IEEE conference on computer vision and pattern recognition}, pages 770--778, 2016.

\bibitem{lang2019pointpillars}
Alex~H Lang, Sourabh Vora, Holger Caesar, Lubing Zhou, Jiong Yang, and Oscar Beijbom.
\newblock Pointpillars: Fast encoders for object detection from point clouds.
\newblock In {\em Proceedings of the IEEE/CVF conference on computer vision and pattern recognition}, pages 12697--12705, 2019.

\bibitem{10077757}
Jinlong Li, Runsheng Xu, Xinyu Liu, Jin Ma, Zicheng Chi, Jiaqi Ma, and Hongkai Yu.
\newblock Learning for vehicle-to-vehicle cooperative perception under lossy communication.
\newblock {\em IEEE Transactions on Intelligent Vehicles}, 8(4):2650--2660, 2023.

\bibitem{li2022bevdepth}
Yinhao Li, Zheng Ge, Guanyi Yu, Jinrong Yang, Zengran Wang, Yukang Shi, Jianjian Sun, and Zeming Li.
\newblock Bevdepth: Acquisition of reliable depth for multi-view 3d object detection.
\newblock {\em arXiv preprint arXiv:2206.10092}, 2022.

\bibitem{li2022bevformer}
Zhiqi Li, Wenhai Wang, Hongyang Li, Enze Xie, Chonghao Sima, Tong Lu, Qiao Yu, and Jifeng Dai.
\newblock Bevformer: Learning bird's-eye-view representation from multi-camera images via spatiotemporal transformers.
\newblock {\em arXiv preprint arXiv:2203.17270}, 2022.

\bibitem{lin2017focal}
Tsung-Yi Lin, Priya Goyal, Ross Girshick, Kaiming He, and Piotr Doll{\'a}r.
\newblock Focal loss for dense object detection.
\newblock In {\em Proceedings of the IEEE international conference on computer vision}, pages 2980--2988, 2017.

\bibitem{liu2021yolostereo3d}
Yuxuan Liu, Lujia Wang, and Ming Liu.
\newblock Yolostereo3d: A step back to 2d for efficient stereo 3d detection.
\newblock In {\em 2021 IEEE International Conference on Robotics and Automation (ICRA)}, pages 13018--13024. IEEE, 2021.

\bibitem{liu2021ground}
Yuxuan Liu, Yuan Yixuan, and Ming Liu.
\newblock Ground-aware monocular 3d object detection for autonomous driving.
\newblock {\em IEEE Robotics and Automation Letters}, 6(2):919--926, 2021.

\bibitem{liu2020smoke}
Zechen Liu, Zizhang Wu, and Roland T{\'o}th.
\newblock Smoke: Single-stage monocular 3d object detection via keypoint estimation.
\newblock In {\em Proceedings of the IEEE/CVF Conference on Computer Vision and Pattern Recognition Workshops}, pages 996--997, 2020.

\bibitem{liu2021autoshape}
Zongdai Liu, Dingfu Zhou, Feixiang Lu, Jin Fang, and Liangjun Zhang.
\newblock Autoshape: Real-time shape-aware monocular 3d object detection.
\newblock In {\em Proceedings of the IEEE/CVF International Conference on Computer Vision}, pages 15641--15650, 2021.

\bibitem{lu2021geometry}
Yan Lu, Xinzhu Ma, Lei Yang, Tianzhu Zhang, Yating Liu, Qi Chu, Junjie Yan, and Wanli Ouyang.
\newblock Geometry uncertainty projection network for monocular 3d object detection.
\newblock In {\em Proceedings of the IEEE/CVF International Conference on Computer Vision}, pages 3111--3121, 2021.

\bibitem{ma2020rethinking}
Xinzhu Ma, Shinan Liu, Zhiyi Xia, Hongwen Zhang, Xingyu Zeng, and Wanli Ouyang.
\newblock Rethinking pseudo-lidar representation.
\newblock In {\em Computer Vision--ECCV 2020: 16th European Conference, Glasgow, UK, August 23--28, 2020, Proceedings, Part XIII 16}, pages 311--327. Springer, 2020.

\bibitem{ma2021delving}
Xinzhu Ma, Yinmin Zhang, Dan Xu, Dongzhan Zhou, Shuai Yi, Haojie Li, and Wanli Ouyang.
\newblock Delving into localization errors for monocular 3d object detection.
\newblock In {\em Proceedings of the IEEE/CVF Conference on Computer Vision and Pattern Recognition}, pages 4721--4730, 2021.

\bibitem{manhardt2019roi}
Fabian Manhardt, Wadim Kehl, and Adrien Gaidon.
\newblock Roi-10d: Monocular lifting of 2d detection to 6d pose and metric shape.
\newblock In {\em Proceedings of the IEEE/CVF Conference on Computer Vision and Pattern Recognition}, pages 2069--2078, 2019.

\bibitem{qin2022monoground}
Zequn Qin and Xi Li.
\newblock Monoground: Detecting monocular 3d objects from the ground.
\newblock In {\em Proceedings of the IEEE/CVF Conference on Computer Vision and Pattern Recognition}, pages 3793--3802, 2022.

\bibitem{reading2021categorical}
Cody Reading, Ali Harakeh, Julia Chae, and Steven~L Waslander.
\newblock Categorical depth distribution network for monocular 3d object detection.
\newblock In {\em Proceedings of the IEEE/CVF Conference on Computer Vision and Pattern Recognition}, pages 8555--8564, 2021.

\bibitem{rukhovich2022imvoxelnet}
Danila Rukhovich, Anna Vorontsova, and Anton Konushin.
\newblock Imvoxelnet: Image to voxels projection for monocular and multi-view general-purpose 3d object detection.
\newblock In {\em Proceedings of the IEEE/CVF Winter Conference on Applications of Computer Vision}, pages 2397--2406, 2022.

\bibitem{simonelli2019disentangling}
Andrea Simonelli, Samuel~Rota Bulo, Lorenzo Porzi, Manuel L{\'o}pez-Antequera, and Peter Kontschieder.
\newblock Disentangling monocular 3d object detection.
\newblock In {\em Proceedings of the IEEE/CVF International Conference on Computer Vision}, pages 1991--1999, 2019.

\bibitem{sindagi2019mvx}
Vishwanath~A Sindagi, Yin Zhou, and Oncel Tuzel.
\newblock Mvx-net: Multimodal voxelnet for 3d object detection.
\newblock In {\em 2019 International Conference on Robotics and Automation (ICRA)}, pages 7276--7282. IEEE, 2019.

\bibitem{song2023graphalign++}
Ziying Song, Caiyan Jia, Lei Yang, Haiyue Wei, and Lin Liu.
\newblock Graphalign++: An accurate feature alignment by graph matching for multi-modal 3d object detection.
\newblock {\em IEEE Transactions on Circuits and Systems for Video Technology}, 2023.

\bibitem{song2023vp}
Ziying Song, Haiyue Wei, Caiyan Jia, Yongchao Xia, Xiaokun Li, and Chao Zhang.
\newblock Vp-net: Voxels as points for 3d object detection.
\newblock {\em IEEE Transactions on Geoscience and Remote Sensing}, 2023.

\bibitem{tian2019fcos}
Zhi Tian, Chunhua Shen, Hao Chen, and Tong He.
\newblock Fcos: Fully convolutional one-stage object detection.
\newblock In {\em Proceedings of the IEEE/CVF international conference on computer vision}, pages 9627--9636, 2019.

\bibitem{10122468}
J. Wang, Y. Zeng, and Y. Gong.
\newblock Collaborative 3d object detection for autonomous vehicles via learnable communications.
\newblock {\em IEEE Transactions on Intelligent Transportation Systems}, 24(9):9804--9816, 2023.

\bibitem{wang2021fcos3d}
Tai Wang, Xinge Zhu, Jiangmiao Pang, and Dahua Lin.
\newblock Fcos3d: Fully convolutional one-stage monocular 3d object detection.
\newblock In {\em Proceedings of the IEEE/CVF International Conference on Computer Vision}, pages 913--922, 2021.

\bibitem{wang2019pseudo}
Yan Wang, Wei-Lun Chao, Divyansh Garg, Bharath Hariharan, Mark Campbell, and Kilian~Q Weinberger.
\newblock Pseudo-lidar from visual depth estimation: Bridging the gap in 3d object detection for autonomous driving.
\newblock In {\em Proceedings of the IEEE/CVF Conference on Computer Vision and Pattern Recognition}, pages 8445--8453, 2019.

\bibitem{yan2018second}
Yan Yan, Yuxing Mao, and Bo Li.
\newblock Second: Sparsely embedded convolutional detection.
\newblock {\em Sensors}, 18(10):3337, 2018.

\bibitem{yang2023bevheight_plus}
Lei Yang, Kaicheng Yu, Tao Tang, Jun Li, Kun Yuan, Li Wang, Yi Huang, Xinyu Zhang, and Peng Chen.
\newblock Bevheight++: Toward robust visual centric 3d object detection.
\newblock {\em arXiv preprint arXiv:2309.16179}, 2023.

\bibitem{yang2023bevheight}
Lei Yang, Kaicheng Yu, Tao Tang, Jun Li, Kun Yuan, Li Wang, Xinyu Zhang, and Peng Chen.
\newblock Bevheight: A robust framework for vision-based roadside 3d object detection.
\newblock In {\em Proceedings of the IEEE/CVF Conference on Computer Vision and Pattern Recognition}, pages 21611--21620, 2023.

\bibitem{yang2023mix}
Lei Yang, Xinyu Zhang, Jun Li, Li Wang, Minghan Zhu, Chuang Zhang, and Huaping Liu.
\newblock Mix-teaching: A simple, unified and effective semi-supervised learning framework for monocular 3d object detection.
\newblock {\em IEEE Transactions on Circuits and Systems for Video Technology}, 2023.

\bibitem{yang2023lite}
Lei Yang, Xinyu Zhang, Jun Li, Li Wang, Minghan Zhu, and Lei Zhu.
\newblock Lite-fpn for keypoint-based monocular 3d object detection.
\newblock {\em Knowledge-Based Systems}, 271:110517, 2023.

\bibitem{ye2022rope3d}
Xiaoqing Ye, Mao Shu, Hanyu Li, Yifeng Shi, Yingying Li, Guangjie Wang, Xiao Tan, and Errui Ding.
\newblock Rope3d: The roadside perception dataset for autonomous driving and monocular 3d object detection task.
\newblock In {\em Proceedings of the IEEE/CVF Conference on Computer Vision and Pattern Recognition}, pages 21341--21350, 2022.

\bibitem{yu2022dair}
Haibao Yu, Yizhen Luo, Mao Shu, Yiyi Huo, Zebang Yang, Yifeng Shi, Zhenglong Guo, Hanyu Li, Xing Hu, Jirui Yuan, et~al.
\newblock Dair-v2x: A large-scale dataset for vehicle-infrastructure cooperative 3d object detection.
\newblock In {\em Proceedings of the IEEE/CVF Conference on Computer Vision and Pattern Recognition}, pages 21361--21370, 2022.

\bibitem{yu2023vehicle}
Haibao Yu, Yingjuan Tang, Enze Xie, Jilei Mao, Jirui Yuan, Ping Luo, and Zaiqing Nie.
\newblock Vehicle-infrastructure cooperative 3d object detection via feature flow prediction.
\newblock {\em arXiv preprint arXiv:2303.10552}, 2023.

\bibitem{yu2023v2x}
Haibao Yu, Wenxian Yang, Hongzhi Ruan, Zhenwei Yang, Yingjuan Tang, Xu Gao, Xin Hao, Yifeng Shi, Yifeng Pan, Ning Sun, et~al.
\newblock V2x-seq: A large-scale sequential dataset for vehicle-infrastructure cooperative perception and forecasting.
\newblock In {\em Proceedings of the IEEE/CVF Conference on Computer Vision and Pattern Recognition}, pages 5486--5495, 2023.

\bibitem{yu2022benchmarking}
Kaicheng Yu, Tang Tao, Hongwei Xie, Zhiwei Lin, Zhongwei Wu, Zhongyu Xia, Tingting Liang, Haiyang Sun, Jiong Deng, Dayang Hao, et~al.
\newblock Benchmarking the robustness of lidar-camera fusion for 3d object detection.
\newblock {\em arXiv preprint arXiv:2205.14951}, 2022.

\bibitem{zhang2022monodetr}
Renrui Zhang, Han Qiu, Tai Wang, Ziyu Guo, Xuanzhuo Xu, Yu Qiao, Peng Gao, and Hongsheng Li.
\newblock Monodetr: depth-guided transformer for monocular 3d object detection.
\newblock {\em arXiv preprint arXiv:2203.13310}, 2022.

\bibitem{zhang2021objects}
Yunpeng Zhang, Jiwen Lu, and Jie Zhou.
\newblock Objects are different: Flexible monocular 3d object detection.
\newblock In {\em Proceedings of the IEEE/CVF Conference on Computer Vision and Pattern Recognition}, pages 3289--3298, 2021.

\bibitem{9729810}
Dingfu Zhou, Xibin Song, Jin Fang, Yuchao Dai, Hongdong Li, and Liangjun Zhang.
\newblock Context-aware 3d object detection from a single image in autonomous driving.
\newblock {\em IEEE Transactions on Intelligent Transportation Systems}, 23(10):18568--18580, 2022.

\bibitem{zhou2019objects}
Xingyi Zhou, Dequan Wang, and Philipp Kr{\"a}henb{\"u}hl.
\newblock Objects as points.
\newblock {\em arXiv preprint arXiv:1904.07850}, 2019.

\bibitem{zhou2021monoef}
Yunsong Zhou, Yuan He, Hongzi Zhu, Cheng Wang, Hongyang Li, and Qinhong Jiang.
\newblock Monoef: Extrinsic parameter free monocular 3d object detection.
\newblock {\em IEEE Transactions on Pattern Analysis and Machine Intelligence}, 44(12):10114--10128, 2021.

\bibitem{zhou2022mogde}
Yunsong Zhou, Quan Liu, Hongzi Zhu, Yunzhe Li, Shan Chang, and Minyi Guo.
\newblock Mogde: Boosting mobile monocular 3d object detection with ground depth estimation.
\newblock {\em Advances in Neural Information Processing Systems}, 35:2033--2045, 2022.

\end{thebibliography}
}

\end{document}